\documentclass[10pt]{article} 
\usepackage[accepted]{tmlr}

\usepackage{amsmath,amsfonts,bm}

\def\eqref#1{equation~\ref{#1}}

\def\1{\bm{1}}

\def\vs{{\bm{s}}}

\DeclareMathAlphabet{\mathsfit}{\encodingdefault}{\sfdefault}{m}{sl}
\SetMathAlphabet{\mathsfit}{bold}{\encodingdefault}{\sfdefault}{bx}{n}

\usepackage[textsize=tiny]{todonotes}
\usepackage{listings}
\usepackage{booktabs} %
\usepackage{glossaries}
\usepackage{algorithm}
\usepackage{pifont}%
\usepackage{xspace}
\usepackage{framed}
\usepackage{color, colortbl}
\usepackage{graphics}
\usepackage{multirow}
\usepackage{caption}
\usepackage{wrapfig}
\usepackage{subfig}
\usepackage{comment}
\usepackage{hyperref}
\usepackage{xcolor}
\definecolor{citecolor}{HTML}{0071BC}
\hypersetup{colorlinks,linkcolor={red},citecolor={citecolor}}
\usepackage{wrapfig}
\usepackage{url}            
\usepackage{booktabs}       
\usepackage{amsfonts}       
\usepackage{nicefrac}       
\usepackage{microtype}      
\usepackage{overpic}        
\newcommand{\my}[1]{\textcolor{red}{[Mingyu: #1]}}

\usepackage{makecell}
\newcommand{\tabincell}[2]{\begin{tabular}{@{}#1@{}}#2\end{tabular}} 

\newlength\savewidth
\newcommand\shline{\noalign{\global\savewidth\arrayrulewidth
  \global\arrayrulewidth 1pt}\hline\noalign{\global\arrayrulewidth\savewidth}}

\makeatletter
\DeclareRobustCommand\onedot{\futurelet\@let@token\@onedot}
\def\@onedot{\ifx\@let@token.\else.\null\fi\xspace}
\def\eg{{e.g}\onedot} 
\def\ie{{i.e}\onedot} 
 
 \def\vs{\emph{vs}\onedot}

\makeatother

\title{Understanding Self-Supervised Pretraining with Part-Aware Representation Learning}

\author{\name Jie Zhu$^{1,2}$\thanks{Equal contribution.}, Jiyang Qi$^{5*}$, Mingyu Ding$^{4*}$, Xiaokang Chen$^3$, \\
        \name Ping Luo$^6$, Xinggang Wang$^5$, Wenyu Liu$^5$, Leye Wang$^{1,2}$, Jingdong Wang$^7$\thanks{Corresponding author.} \\
        \email zhujie@stu.pku.edu.cn,  \{jiyangqi, xgwang, liuwy\}@hust.edu.cn, myding@berkeley.edu, \\
         \{pkucxk, leyewang\}@pku.edu.cn, pluo@cs.hku.hk,  wangjingdong@outlook.com\\
      \addr $^1$Key Lab of High Confidence Software Technologies (Peking University), Ministry of Education, China \\ 
      $^2$School of Computer Science, Peking University, $^3$School of AI, Peking University, $^4$UC Berkeley \\
      $^5$School of EIC, Huazhong University of Science \& Technology, $^6$University of Hong Kong, 
       $^7$Baidu \\}

\def\month{07}  
\def\year{2023} 
\def\openreview{\url{https://openreview.net/forum?id=HP7Qpui5YE}} 

\begin{document}

\maketitle

\begin{abstract}
In this paper, we are interested in understanding self-supervised pretraining through studying the capability that self-supervised methods learn part-aware representations. The study is mainly motivated by that random views, used in contrastive learning, and random masked (visible) patches, used in masked image modeling, are often about object parts.

We explain that contrastive learning is a~\emph{part-to-whole} task: the projection layer hallucinates the whole object representation from the object part representation learned from the encoder, and that masked image modeling is a \emph{part-to-part} task: the masked patches of the object are hallucinated from the visible patches. The explanation suggests that the self-supervised pretrained encoder leans toward understanding the object part. We empirically compare the off-the-shelf encoders pretrained with several representative methods on object-level recognition and part-level recognition. The results show that the fully-supervised model outperforms self-supervised models for object-level recognition, and most self-supervised contrastive learning and masked image modeling methods outperform the fully-supervised method for part-level recognition. It is observed that the combination of contrastive learning and masked image modeling further improves the performance.
\end{abstract}

\section{Introduction}
Self-supervised representation pretraining
has been attracting a lot of research efforts recently.
The goal is to train an encoder
that maps an image to a representation
from visual contents without the necessity 
of human annotation. Through this, it is expected that the encoder benefits the downstream tasks,
\eg, segmentation and detection.

\begin{figure}[t]
    \centering
    \footnotesize
    \subfloat[]{\includegraphics[height=0.172\linewidth]{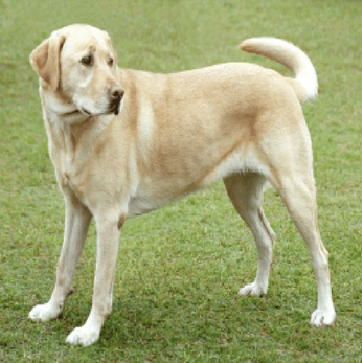}\label{fig:crop_1}}\hfill
    \subfloat[]{\includegraphics[height=0.172\linewidth]{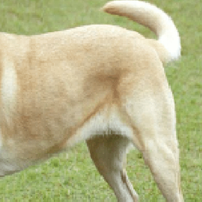}\label{fig:crop_2}}\hfill
    \subfloat[]{\includegraphics[height=0.172\linewidth]{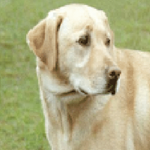}\label{fig:crop_3}}\hfill
    \subfloat[]{\includegraphics[height=0.172\linewidth]{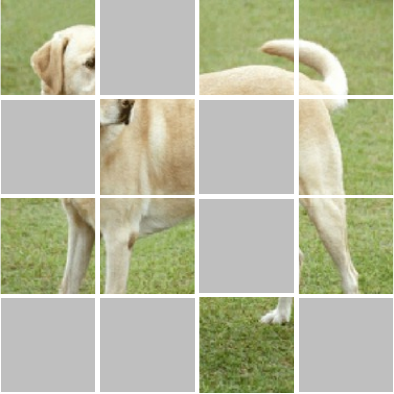}\label{fig:crop_4}}\hfill
    \subfloat[]{\includegraphics[height=0.172\linewidth]{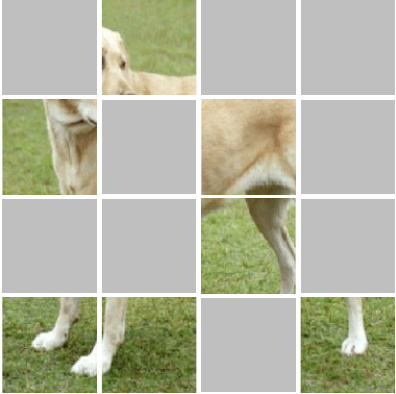}\label{fig:crop_5}}
    \vspace{-8pt}
    \caption{(a) original image,
    (b-c) two random crops,
    and (d-e) masked and visible patches.}
    \label{fig:randomcropsdog}
    \vspace{-12pt}
\end{figure}

There are two main frameworks:
contrastive learning\footnote{In this paper, we use contrastive learning (CL)
to refer to random-view based methods,
\eg, SimCLR, MoCo, and BYOL.
}
and masked image modeling.
Contrastive learning
aims to
maximize the agreement 
of the embeddings
of random augmented views 
from the same image.
Masked image modeling 
partitions an image
into masked patches and visible patches,
and makes predictions
for masked patches from visible patches.
Figure~\ref{fig:randomcropsdog} gives examples
of random views for contrastive learning
and masked and visible patches 
for masked image modeling.

We observe that
a random view and
a set of masked (visible) patches
usually contain a portion of an object.
It is also reported in self-supervised learning methods, \eg, DINO~\citep{caron2021emerging} and iBOT~\citep{zhou2021ibot},
that different attention heads in ViTs can attend to different semantic regions or parts of an object.
In light of this, we attempt to understand self-supervised pretraining
by studying the capability that
the pretrained encoder
learns part representations.

\begin{figure*}[t]
\footnotesize
    \hspace{3.05cm}DeiT\hspace{3.55cm}MoCo v3\hspace{3.85cm}CAE
    \centering
    \includegraphics[width=.98\linewidth]{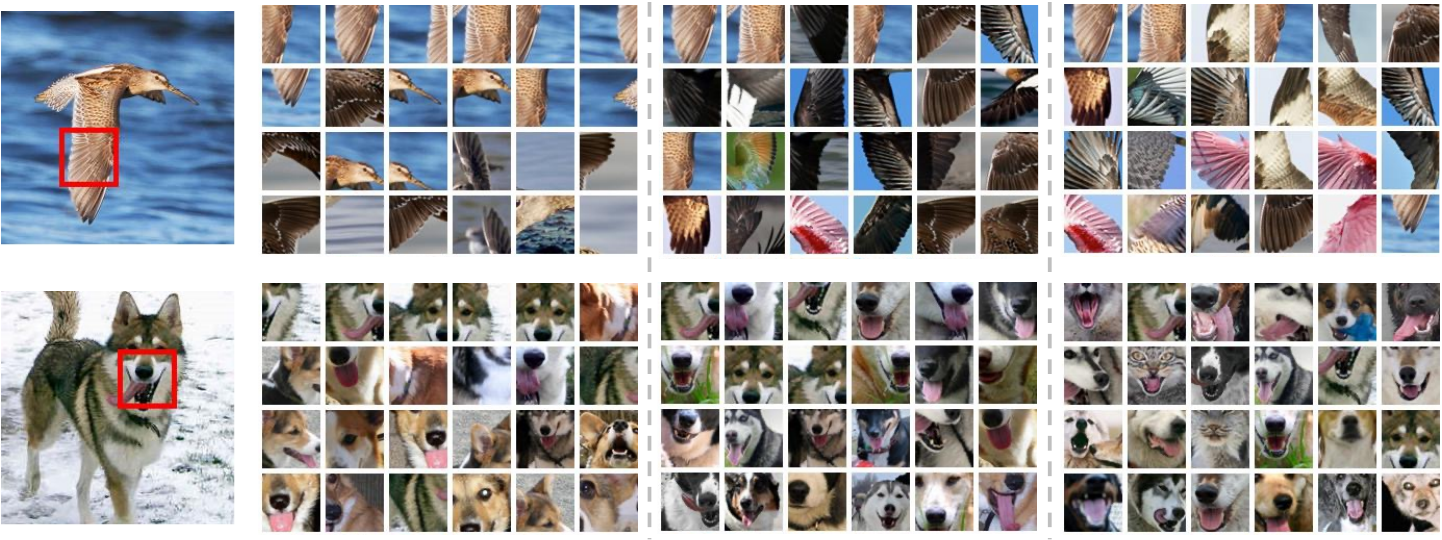}
    \vskip -0.1in
    \caption{Top-24 patch retrieval results 
    with three frozen encoders of 
    DeiT, MoCo v3, and CAE,
    by taking the patch in the red box as the query.
    It can be seen that
    the retrieved results from CAE and MoCo v3 
    are about the object part (wing and dog mouth)
    and more precise than DeiT (about the whole object)
    implying that self-supervised pretraining methods, CAE and MoCo v3 are stronger at learning part-aware representations than the fully-supervised method DeiT.
    Details could be found in Sec.~\ref{sec:method}.}
    \vspace{-4pt}
    \label{fig:searchresultsteaser}
    \vskip -0.1in
\end{figure*}

We present a~\emph{part-to-whole} explanation
for typical contrastive learning methods
(\eg, SimCLR~\citep{ChenK0H20}, MoCo~\citep{mocov3_chen2021empirical}, and BYOL~\citep{grill2020bootstrap}):
the embedding of the whole object
is hallucinated from the embedding
of the part of the object contained 
in the random crop
through a projection layer.
In this way, embeddings of random crops from the same image naturally agrees with each other.
Masked image modeling is a~\emph{part-to-part} process:
the embeddings of the masked patches of the object
(a part of the object),
are hallucinated
from the visible patches
(the other part of the object).

We empirically compare
the supervised model DeiT~\citep{touvron2020deit}, which serves as an important baseline for analyzing SSL models, and 
typical self-supervised representation pretraining methods, including MoCo v3~\citep{mocov3_chen2021empirical}, DINO~\citep{caron2021emerging}, CAE~\citep{cae_chen2022context}, MAE~\citep{he2021masked}, BEiT~\citep{bao2021beit}, and iBOT~\citep{zhou2021ibot},
on object-level recognition (image classification and object segmentation) and part-level recognition
(patch retrieval, patch classification, and part segmentation).
Figure~\ref{fig:searchresultsteaser}
presents patch retrieval results
using the encoders learned
through CAE, MoCo v3, and DeiT,
implying that the encoders 
pretrained by CAE and MoCo v3 are able to learn
part-aware representations.

Through extensive studies 
and comparisons, 
we make the following observations.
1) DeiT outperforms 
contrastive learning and MIM 
methods except iBOT in object-level recognition tasks, which may benefit from its explicit object-level supervision.
2) In contrast, self-supervised methods learn better part-aware representations than DeiT.
For example, while DeiT is superior to DINO and CAE by 0.4\% and 2.3\% on ADE20K object segmentation, DINO and CAE outperform DeiT by 1.6\% and 1.1\% on ADE20K part segmentation, respectively.
3) In contrastive learning, the encoder can learn part-aware information, while the projected representation tends to be more about the whole object. The evidence could be found in part retrieval experiments
on MoCo v3, DINO, and iBOT.
4) The MIM method CAE shows good potential in part-aware representation learning. Interestingly, the method combines contrastive learning and MIM is promising, \eg, iBOT learns better representations at both object and part levels.

%

This paper presents the following contributions:
\begin{itemize}
\setlength\itemsep{0.03 em}
    \vspace{-7pt}
    \item We 
    study the capability of learning part-aware representations
    as a way of 
    understanding self-supervised representation pertaining based on both qualitative analysis and empirical results.
    \item We explain masked image modeling as a part-to-part task
    and contrastive learning as a part-to-whole task,
    and speculate that self-supervised pretraining has the potential
    for learning part-aware representations.
    \item We empirically compare several pretrained models on object-level and part-level recognition tasks, showing interesting findings with supporting evidence of the capability of part-aware representation learning for self-supervised learning. Code and dataset are available~\footnote{\url{https://github.com/Atten4Vis/} and \url{https://github.com/JiePKU/understand-ssl-part-aware}}.
\end{itemize}

\section{Related Work}

\begin{figure*}[t]
\centering
\centerline{\includegraphics[scale=0.7]{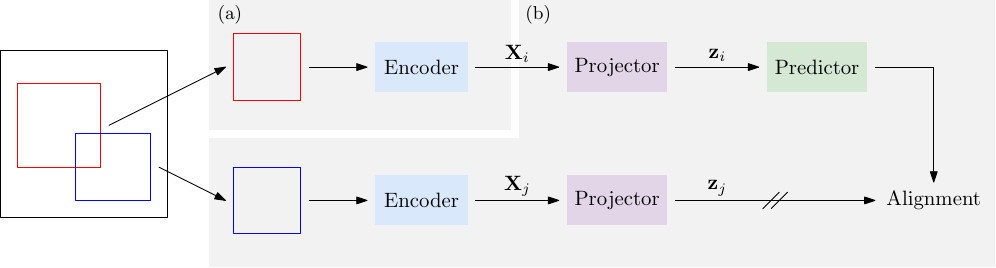}}
\vspace{-5pt}
\caption{The pipeline of a typical contrastive learning approach.
Two augmented views,
red box and blue box,
are generated from the original image.
The augmented view in red is fed into
the encoder and the projector,
and then the predictor (which does not appear in earlier works like MoCo~\citep{mocov3_chen2021empirical} and SimCLR~\citep{ChenK0H20}),
and the view in blue is fed into 
the encoder and the projector.
The two outputs are expected to be aligned.
The gradient is stopped for the bottom stream.
}
\vspace{-12pt}
\label{fig:CP} 
\end{figure*}

\noindent\textbf{Contrastive learning (CL).} 
Contrastive pretraining has been an intense academic field in the CNN era.
In this work, we use it to refer to methods for comparing random views~\citep{caron2020unsupervised_swav,ChenK0H20,zbontar2021barlow,ssl_swin_xie2021self,mocov3_chen2021empirical,caron2021emerging}, including some instance discrimination work such as \citep{grill2020bootstrap,chen2021exploring,bardes2021vicreg}.
A representative work, SimCLR~\citep{ChenK0H20} learns representations through maximizing agreement between different views of the same image in the latent space. BYOL~\citep{grill2020bootstrap} uses two asymmetrical networks to bootstrap latent representation without negative samples involved during the interaction. As vision transformer (ViT)~\citep{DosovitskiyB0WZ21} shows excellent performance via supervised learning, it is adopted subsequently in contrastive pertaining, and numerous outstanding works are proposed. For example, MoCo v3~\citep{mocov3_chen2021empirical} observes the hidden instability while training self-supervised ViT and solves it by using a fixed random patch projection. DINO~\citep{caron2021emerging} explores new properties derived from self-supervised ViT and accordingly designs a learning strategy interpreted as a form of self-distillation with no labels.

\noindent\textbf{Masked image modeling (MIM).} Masked image modeling is another self-supervised pretraining paradigm that attracts much attention recently.
BEiT~\citep{bao2021beit} follows masked language modeling in the natural language process (NLP) area and predicts tokens via mapping image patches by d-VAE~\citep{d-vae_ramesh2021zero}. PeCo~\citep{dong2021peco} boosts BEiT by taking into consideration more semantic information in visual tokens. MAE~\citep{he2021masked} learns rich hidden information by directly performing masked image reconstruction in RGB color space using ViT while SimMIM~\citep{xie2021simmim} uses Swin-transformer~\citep{swin_liu2021swin}. CAE~\citep{cae_chen2022context} adds a regressor between encoder and decoder, which is designed to align unmasked patches with masked ones, leading to a pure context encoder. Recently, a trend that combines MIM with siamese frameworks has surfaced and showed encouraging results including MST~\citep{mst_li2021mst}, SplitMask~\citep{el2021large}, iBOT~\citep{zhou2021ibot}, dBOT~\citep{liu2022exploring}, and SIM~\citep{tao2022siamese}.

\noindent\textbf{Understanding self-supervised contrastive pretraining.}
The studies on understanding contrastive pretraining~\citep{SaunshiAGMZAKK22, chen2022intra, zhong2022self, wei2022contrastive}
mainly focus on 
random augmentations (views),
contrastive loss function and its variants
under the assumption that: 
the augmentations of inputs from the same class have significant overlap
in the representation space, 
but there is little overlap for inputs from different classes.
Our work is complementary to these studies.
Inspired by the observation that random views usually contain a portion of an object, and methods~\citep{caron2021emerging,zhou2021ibot} show that different attention heads in ViTs can attend to different semantic regions of an object, we investigate what the encoder and the projector do in typical self-supervised contrastive pretraining.
We speculate that the pretraining task is a part-to-whole problem, 
predicting the representation of the whole object
through the projector
from the representation (obtained from the encoder) of the part of an object.
We use empirical results to verify our analysis.

\noindent\textbf{Understanding self-supervised masked image modeling.}
The comparison of attention in different layers 
between the pretrained models from MIM and the supervised approach 
is conducted:
MIM pretraining brings locality to the trained model with sufficient diversity on the attention heads~\citep{XieGHZHC22}.
Consistent with the analysis in 
NLP, empirical studies are conducted in~\cite{XieZGLWDH22}
to verify that
MIM benefits from 
larger models, more data,
and longer training. 
CAE~\citep{cae_chen2022context} gives the comparison
between contrastive and MIM and shows MIM cares about all patches and thus achieves better results for fine-tuning.
\cite{cao2022understand} provides a mathematical understanding of MIM. \cite{kong2022understanding} points out that the learned occlusion invariant feature contributes to the success of MIM.
In this work, we speculate that masked image modeling is a~\emph{part-to-part} process: the embeddings of the masked part of the object are hallucinated from the visible part using the position information of the masked patches, leading to better part-aware representation than the supervised model DeiT~\citep{touvron2020deit}. More discussions can be found in Appendix~\ref{app:discussion}.

\section{Understanding Contrastive Learning
and Mask Image Modeling}
\label{sec:method}
In this section, we briefly review representative formulations of CL and MIM, and provide qualitative analysis of their part-aware explanations.
\subsection{Contrastive Learning}

Contrastive learning
aims to learn the encoder 
through maximizing the agreement 
between differently augmented views of the same image
in the representation space.
An example pipeline is depicted in Figure~\ref{fig:CP}.
Given an image $\mathsf{I}$,
the augmentations,
\eg, random cropping,
random color distortion,
and random Gaussian blur,
are applied
to generate a set of $N$ augmented views,
$\{\mathsf{V}_1,
\mathsf{V}_2,
\cdots,
\mathsf{V}_N\}$.
An augmented view $\mathsf{V}_n$
is fed into
an encoder $\operatorname{Encoder}$,
generating the encoded representation $\mathbf{x}_n$,
and followed by a projector,
generating the projection $\mathbf{z}_n$.
The basic goal is 
to maximize the agreement between the projections
$\{\mathbf{z}_1, \mathbf{z}_2, \cdots, \mathbf{z}_N\}$,
\ie, minimize the loss 
\begin{equation}
\begin{split}
    \mathcal{L}_{\operatorname{CPT}} = \sum\nolimits_{i=1}^N\sum\nolimits_{j=1}^N 
    \ell\{\operatorname{Projector}(\operatorname{Encoder}(\mathsf{V}_i)),  \operatorname{Projector}(\operatorname{Encoder}(\mathsf{V}_j))\}.
    \label{eqn:CPTask}
\end{split}
\end{equation}
In the formulation with a contrastive loss,
the agreement between the projections of random augmentations
from different images is minimized.

\noindent\textbf{Part-to-whole prediction explanation.} 
Let us consider two crops 
randomly sampled from the original image 
(see the examples given in Figure~\ref{fig:randomcropsdog}(b-c)).
The encoded representation of the first crop
is expected to describe a part of the object dog;
the encoded representation of the second crop
is expected to describe another part of the object dog\footnote{In this paper, we mainly study the capability of learning
representations about objects and parts,
and leave the study of representations
of the background as the future work.}.
The two representations are related but different.
Contrastive learning methods project the two encoded representations
into two projected representations
that are expected to agree.
We hypothesize that
\emph{the projection process
maps the encoded part representation
to the representation of the whole object}\footnote{It
is said that different parts
have common causes in the external world~\citep{BeckerH92}.
Our hypothesis is 
that the common cause is the whole object.}.
Through this way,
the projected representations will agree
to different views
from the same image~\footnote{For methods that use local-global crops, \eg, DINO, contains a range of $0.05-0.4$ for local crops and $0.4-1$ for global crops, which in our hypothesis are expected to describe \textit{smaller} and \textit{larger} parts of the object, respectively.}.
It is assumed that the part-to-whole projection is more reliable if the encoded representation is semantically richer
and is able to describe the part information.
The part-to-whole process suggests 
that the encoder pretrained by contrastive learning methods
is potentially capable of learning part-aware representations.

\begin{figure}
\footnotesize
    \hspace{3.7cm} \footnotesize Encoded representation \hspace{2.5cm} \footnotesize Projected representation
    \centering
    \includegraphics[width=.9\linewidth]{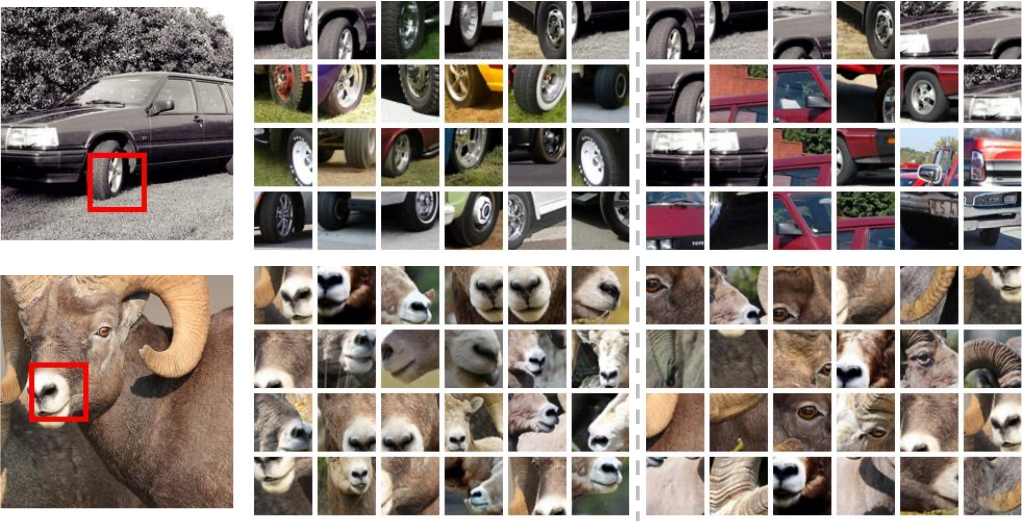}
    \vskip -0.1in
    \caption{Illustration
    of patch search results 
    using encoded representations
    and projections
    (pretrained with MoCo v3 as).
    Left: patch search results 
    with encoded representations.
    Right: patch search results 
    with projections.
    In each result,
    the small patch encircled by the red box is taken as the query.
    It can be seen
    that for encoded representations,
    the returned patches
    are about the same part,
    and for projections,
    the result patches
    are about the same object,
    verifying the \emph{part-to-whole} hypothesis.
    }
    \label{fig:part2whole}
    \vspace{-2pt}
    \vskip -0.1in
\end{figure}

Figure~\ref{fig:part2whole} provides patch search results of a representative contrastive learning method MoCo v3~\citep{mocov3_chen2021empirical}
based on the encoded representations before and after
the projections. 
The visualized patch retrieval results in Figure~\ref{fig:part2whole} as well as those in Figure~\ref{fig:searchresultsteaser} are obtained based on ImageNet~\citep{deng2009imagenet} validation set. Concretely, we uniformly crop 49 patches sized 56$\times$56 using a stride of 28 from each pre-processed 224$\times$224 validation image in ImageNet. With all the cropped patches from the validation set, we select one patch as a query and find the top 24 patches with the highest cosine similarity with it.

One can see Figure~\ref{fig:part2whole} that the results 
through the encoded representations 
are mainly about the local part,
and the results through the projections
tend to include the other parts of the same object.
In other words,
the projections tend
to be about the whole object. The search results 
verify the part-to-whole hypothesis. Similar observations are also shown in \cite{chen2022intra} where the embedding space tends to preserve more equivariance and locality, while the projection has more invariance.

\subsection{Masked Image Modeling}

Mask image modeling is the task of predicting some parts of an image from the remaining parts.
An augmented view of an image is partitioned into patches,
$\mathcal{R} = \{\mathsf{R}_1, \mathsf{R}_2, \dots,
\mathsf{R}_M\}$.
The task is to predict a subset of patches 
$\mathcal{R}_m$,
named masked patches,
from the remaining patches $\mathcal{R}_v$, 
named visible patches.
%
Considering contrastive learning that 
explicitly compares representations of
random views, 
we take context autoencoder (CAE)~\citep{cae_chen2022context} as an example 
that explicitly predicts 
the encoded representations of the masked patches
from the encoded representations of the visible patches\footnote{Some
MIM methods,
such as BEiT~\citep{bao2021beit} and MAE (Masked Autoencoder)~\citep{he2021masked}
do not have an explicit process 
to predict the encoded representations of masked patches,
instead,
directly reconstruct the targets.}.


\begin{figure*}[t]
\centering
\centerline{\includegraphics[scale=0.7]{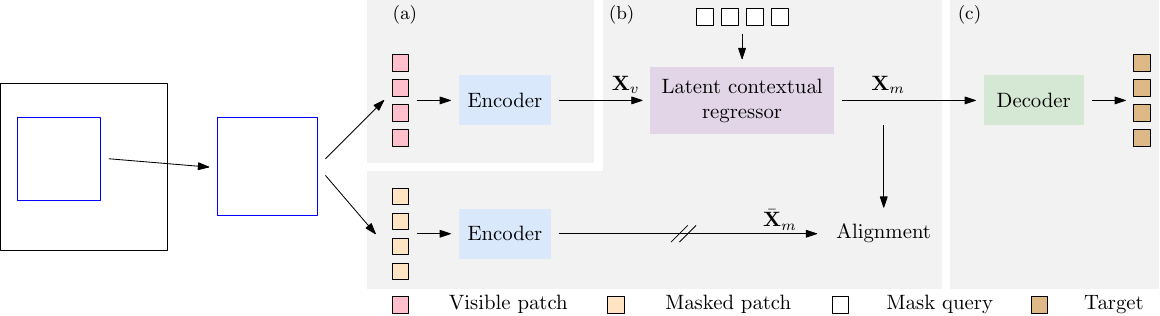}}
\caption{The pipeline of an MIM approach,
context autoencoder (CAE).
An augmented view (in blue) of the image is partitioned into visible and masked patches.
The CAE approach
feeds visible patches into
the encoder
and extracts their representations $\mathbf{X}_v$
and then completes the pretext task
by
predicting
the representations $\mathbf{X}_m$
of the masked patches
from the visible patches
in the encoded representation space
with latent contextual regressor and alignment constraint,
and mapping predicted 
representations $\mathbf{X}_m$
of masked patches
to the targets.
The pretrained encoder in (a) 
is applied to downstream tasks
by simply replacing the pretext task part
(b, c)
with the downstream task completion part.
}
\label{fig:CAE}
\vskip -0.12in
\end{figure*}

One goal of CAE
(illustrated in Figure~\ref{fig:CAE}),
which we call masked representation modeling (MRM),
is to maximize the agreement 
between the predictions of the representations
of masked patches (through a regressor)
and the representation of masked patches computed
from the encoder
by minimizing the loss 
\begin{align}
 \ell_{\operatorname{MRM}}(
\operatorname{Regressor}(\operatorname{Encoder}(\mathcal{R}_v)),
    \operatorname{Encoder}(\mathcal{R}_m)).
\end{align}
Here, we do not include the positional embeddings
of masked and visible patches for clarity.
%
%
It is noted that MRM differs from contrastive learning: 
MRM does not compare multiple random views,
but compares the regressed representations for masked patches
and the encoded representations of masked patches.
In addition,
there is another loss
for target prediction (reconstruction) for the masked patches, which is commonly used in masked image modeling (MIM) methods:
\begin{equation}
    \ell_{\operatorname{MIM}}(
    \operatorname{Decoder}(
\operatorname{Regressor}(\operatorname{Encoder}(\mathcal{R}_v)
)), \operatorname{Target}(\mathcal{R}_m)),
\end{equation}
where $\operatorname{Target}(\mathcal{R}_m)$
is a function to map the masked patches
to the targets,
\eg, d-VAE~\citep{d-vae_ramesh2021zero} token used in CAE and BeiT~\citep{bao2021beit}, or normalized RGB values used in MAE~\citep{he2021masked}.

%
%

%

\noindent\textbf{Part-to-part
prediction explanation.}
The masked image modeling approaches,
including CAE, MAE, and BEiT, 
{make use of
the positions of masked patches}
for making predictions for masked patches
from visible patches.
The visible patches and masked patches
often contain different parts of an object.
In other words,
MIM aims to predict the masked part
of an object from the visible part.
We name this a \emph{part-to-part} process.
There are two part-to-part tasks:
one is to reconstruct the part targets
from the visible part representations (MAE and CAE) or
from the visible part raw pixels (BEiT),
and the other one is
to regress the masked part representations (CAE).
The part-to-part process suggests 
that the encoder pretrained by MIM methods
is potentially capable of learning part-aware representations.
Figure~\ref{fig:searchresultsteaser} illustrates the capability with the patch retrieval results. More visualizations can be found in Appendix~\ref{app:visualization}.


\section{Experiments}

\begin{figure}[h]
\begin{minipage}[t]{0.45\linewidth}
\begin{table}[H]
\centering  
  \caption{
  Top-1 accuracy with linear probing, and attentive probing~\citep{cae_chen2022context}, on the ImageNet classification benchmark~\citep{deng2009imagenet}.
  }
  \vspace{0.1cm}
\setlength{\tabcolsep}{16pt}
\renewcommand{\arraystretch}{1.2}
    \footnotesize 
    \begin{tabular}{l c c}
      \shline
      Method & Linear & Attentive \\
      \shline
      \multicolumn{3}{l}{\emph{Supervised Model}:}\\
      DeiT & 81.8 & 81.8   \\
      \hline
      \multicolumn{3}{l}{\emph{Contrastive Learning}:}\\
      MoCo v3 & 76.2 & 77.0\\
      DINO & 77.3 & 77.8 
      \\
      \hline
      \multicolumn{3}{l}{\emph{Masked Image Modeling (MIM)}:}\\
      BEiT  & 41.8 & 51.9 \\
      MAE  & 67.8 & 74.2 \\
      CAE  & 70.4 & 77.1 \\
      \hline
      \multicolumn{3}{l}{\emph{Contrastive Learning + MIM}:}\\
      iBOT  & 79.5 & 79.8 \\
      \shline
  \end{tabular}
  \label{tab:imagenet_classification}
\end{table}
\end{minipage}
\hfill
\begin{minipage}[t]{0.51\linewidth}
\begin{table}[H]
\setlength{\tabcolsep}{5pt}
\renewcommand{\arraystretch}{1.1}
\centering
\footnotesize
    \caption{Object retrieval results on CIFAR10 and the cropped object of COCO. The ``Encoded" and ``Projected" refer to the encoded
    and projected representations.
    }
    \label{tab:quantitative_object_retrieval}
    \begin{tabular}{l c c c c c c}
    \shline
        \multirow{3}{*}{Methods} && \multicolumn{5}{c}{Object Retrieval (AP)} \\
        \cline{3-7}
        && \multicolumn{2}{c}{CIFAR10} && \multicolumn{2}{c}{COCO} \\
        \cline{3-4} \cline{6-7}
        && Encoded  & Projected && Encoded  & Projected  \\
        \shline
        \multicolumn{4}{l}{\emph{Supervised Model}:}\\
        DeiT && 71.3 & -- && 71.2 & --  \\
        \hline
      \multicolumn{4}{l}{\emph{Contrastive Learning}:}\\
        MoCo v3 && 40.6 &	56.0	&&	47.7 &	55.9  \\
        DINO && 31.3 &	61.5	&&	38.0 &	57.3  \\
        \hline
      \multicolumn{4}{l}{\emph{Masked Image Modeling (MIM)}:}\\
        BEiT && 14.6 & -- && 11.7 & --  \\
        MAE && 15.9 & -- && 13.8 & --  \\
        CAE && 39.1 & -- && 46.0 & --  \\
        \hline
      \multicolumn{4}{l}{\emph{Contrastive Learning + MIM}:}\\
        iBOT && 40.2 &	60.0	&&	53.2	& 55.8 \\
        \shline
    \end{tabular}
\end{table}
\end{minipage}
\end{figure}


We study seven representative methods with the same ViT-B encoder, including a supervised method DeiT~\citep{touvron2020deit} that serves as an important baseline for analyzing SSL models, contrastive learning methods MoCo v3~\citep{mocov3_chen2021empirical}, DINO \citep{caron2021emerging}; masked image modeling (MIM) methods BEiT~\citep{bao2021beit}, MAE~\citep{he2021masked}, and CAE~\citep{cae_chen2022context}; and iBOT~\citep{zhou2021ibot} that combines contrastive learning and MIM.
We take the training epochs specified in each work to ensure that all compared models are properly trained and use publicly available checkpoints~\footnote{
Considering that the training protocols of different models vary widely in most respects (\eg, batch size, augmentation, and ViT-related tricks to prevent training instability and crashes), it is unreasonable to use the identical setup. For example, DINO uses a centering and sharpening strategy to avoid collapse.
}:
300 for DeiT, 300 (600\footnote{The number of effective training epochs introduced in \cite{zhou2021ibot}.\label{ft:effective_epochs}}) for MoCo~v3, 400 (1600\textsuperscript{\ref{ft:effective_epochs}}) for DINO and iBOT, 800 for BEiT, and 1600 for MAE and CAE. Frozen encoders are used in all experiments to understand what these different representation pretraining methods learn. More details about models can be found in Appendix~\ref{app:model_description}.

\subsection{Object-Level Recognition}
\label{sec:object_level}
We benchmark three widely-studied object-level recognition, \ie, image classification, object retrieval, and semantic segmentation
to show the capability
that the pretrained encoder
learns object-level representations.

\noindent \textbf{Image classification.} We report the linear probing, and attentive probing results of the selected models on ImageNet~\citep{deng2009imagenet}. For attentive probing, we follow the protocol in CAE~\citep{cae_chen2022context} that append a cross-attention layer together with a batch normalization
layer and a linear classifier.

We have the following observations from Table~\ref{tab:imagenet_classification}.
\ding{182} The supervised model, DeiT performs better than self-supervised models at object-level recognition.
\ding{183} The models that leverage contrastive learning, \ie, MoCo, DINO, and iBOT, show superior linear probing performance than MIM-based models, demonstrating they contain more object-aware high-level semantics.
\ding{184} MIM-based models, \eg, CAE, show inferior results in linear probing while competitive results with contrastive-based methods in attentive probing. The reason might be that MIM is capable of attending to all the regions, including non-object regions in an image, thus needs a spatial feature selection step to attend to the object part,
which is pointed out in~\cite{cae_chen2022context}.
BEiT and MAE perform inferior in linear and attentive probes, 
implying that the two methods are less capable
of learning semantics.

\noindent \textbf{Object-level retrieval.} We evaluate the object-level retrieval performance on two datasets, \ie, CIFAR10~\citep{krizhevsky2009learning} and COCO~\citep{lin2014microsoft}. For CIFAR10, we directly use each image (resized to $224\times224$) to perform the retrieval task. For COCO, we build the retrieval database by leveraging the bounding box (that is approximately square) in the annotation file to crop the objects and resizing to $224\times224$. The retrieval results are reported in Table~\ref{tab:quantitative_object_retrieval}. One can see that DeiT obtains the best performance over self-supervised methods in both datasets, showing stronger object-aware capability. Besides, we also notice that for part-to-whole SSL methods, \eg, MoCo v3, the projected feature has higher accuracy than the encoded one, possibly because the projection boosts the object-aware representations, which further supports our part-to-whole hypothesis.

\begin{wraptable}{r}{0.45\linewidth}
    \centering
    \footnotesize
    \setlength{\tabcolsep}{12pt}
    \vspace{-12pt}
    \renewcommand{\arraystretch}{1.2}
    \caption{
    Linear evaluation of ADE20K~\citep{ade20k_zhou2019semantic} object-level semantic segmentation (150 classes) using $4\times$ upsampling and a single $1\times 1$ convolutional layer on frozen backbones.
    }
    \label{tab:ADE20K_segmentation}
\begin{tabular}{l c c c}
    \shline
        Method & mIoU & mAcc & aAcc \\
        \shline
        \multicolumn{4}{l}{\emph{Supervised Model}:}\\
        DeiT & 34.9 & 44.2 & 75.4\\
        \hline
      \multicolumn{4}{l}{\emph{Contrastive Learning}:}\\
        MoCo v3 & 34.7 & 43.9 & 75.9 \\
        DINO & 34.5 & 43.5 & 76.1 \\
        \hline
      \multicolumn{4}{l}{\emph{Masked Image Modeling (MIM)}:}\\
        BEiT & 17.8 & 23.7 & 64.9  \\
        MAE & 27.1 & 34.8 & 71.6   \\
        CAE & 32.6 & 42.2 & 75.2  \\
        \hline
      \multicolumn{4}{l}{\emph{Contrastive Learning + MIM}:}\\
        iBOT & 38.3 & 47.4 & 78.1 \\
        \shline
    \end{tabular}
    \vspace{-12pt}
\end{wraptable}  
\noindent \textbf{Object-level semantic segmentation.} We perform linear evaluation on ADE20K~\citep{ade20k_zhou2019semantic} to 
show the object-level semantic capabilities of the pretrained models. A $4\times$ bilinear interpolation and a single $1\times 1$ convolutional layer for pixel labeling are attached to the frozen encoder.
The learning rate ($4e-4$), training iterations ($160k$), and batch size ($16$) among all the experiments maintain the same during training for fair comparisons. The input size is set to $512 \times 512$ following previous works~\citep{bao2021beit, he2021masked, cae_chen2022context, zhou2021ibot}.

We can see from Table~\ref{tab:ADE20K_segmentation}
that the supervised model DeiT outperforms all self-supervised models except iBOT, including  contrastive learning and MIM methods on ADE20K object-level segmentation. 
This implies that these self-supervised models are not strong at object-level understanding,
which is consistent with the observations
for image classification. 
%
%
iBOT~\citep{zhou2021ibot}, as a combination of contrastive learning and MIM, shows surprisingly better performance than the supervised model DeiT on ADE20K,
implying the power
of combining contrastive learning and 
masked image modeling 
for downstream tasks.

\subsection{Part-Level Recognition}

Self-supervised methods like iBOT~\citep{zhou2021ibot} and DINO~\citep{caron2021emerging} qualitatively show that different attention heads in ViTs can attend to different semantic regions of an object.
We conduct the quantitative evaluation for part-aware representation 
obtained by pretrained models that is not well explored before,
through three part-level recognition tasks,
part retrieval, part classification, and part segmentation.

\begin{table*}[h]
\setlength{\tabcolsep}{7pt}
\renewcommand{\arraystretch}{1.1}
\centering
\footnotesize
    \caption{Part retrieval and classification results on the cropped part patches of CUB-200-2011 and COCO. The ``Encoded" and ``Projected" refer to the encoded
    and projected representations. ``Linear" and ``Attentive" columns denote the linear probing and attentive probing accuracy, respectively.
    }
    \label{tab:quantitative_part_retrieval}
    \begin{tabular}{l c c c c c c c c c c c c}
    \shline
        \multirow{3}{*}{Methods} && \multicolumn{5}{c}{Part Retrieval (AP)} && \multicolumn{5}{c}{Part Classification (Acc)} \\
        \cline{3-7} \cline{9-13}
        && \multicolumn{2}{c}{CUB-200-2011} && \multicolumn{2}{c}{COCO} && \multicolumn{2}{c}{CUB-200-2011} && \multicolumn{2}{c}{COCO} \\
        \cline{3-4} \cline{6-7} \cline{9-10} \cline{12-13}
        && Encoded  & Projected && Encoded  & Projected  && Linear & Attentive && Linear & Attentive \\
        \shline
        \multicolumn{4}{l}{\emph{Supervised Model}:}\\
        DeiT && 35.0 & -- && 44.1 & -- && 90.9 & 92.9 && 88.5 & 91.4 \\
        \hline
      \multicolumn{4}{l}{\emph{Contrastive Learning}:}\\
        MoCo v3 && 50.8 & 28.4 && 52.3 & 36.8 && 93.8 & 96.0 && 92.4 & 95.3 \\
        DINO && 48.9 & 31.7 && 51.8 & 41.2 && 93.2 & 95.2 && 91.7 & 94.5 \\
        \hline
      \multicolumn{4}{l}{\emph{Masked Image Modeling (MIM)}:}\\
        BEiT && 27.9 & -- && 35.3 & -- && 55.4 & 86.5 && 69.3 & 86.5 \\
        MAE && 28.5 & -- && 37.1 & -- && 86.9 & 92.8 && 88.0 & 93.9 \\
        CAE && 58.0 & -- && 57.0 & -- && 89.5 & 95.8 && 91.1 & 95.5 \\
        \hline
      \multicolumn{4}{l}{\emph{Contrastive Learning + MIM}:}\\
        iBOT && 49.3 & 31.2 && 59.2 & 41.5 && 93.8 & 95.8 && 92.1 & 95.1 \\
        \shline
    \end{tabular}
\end{table*}

\noindent\textbf{Part retrieval.}
We conduct part retrieval experiments
on two datasets,
CUB-200-2011~\citep{cub200_wah2011caltech} and COCO~\citep{lin2014microsoft}, which provide both the positions and corresponding categories of the keypoints. 
We build the part patch databases
by cropping the patches where these keypoints are located at the center.
We consider four and three keypoints
from the two datasets, respectively.
%
For each keypoint, 
we find the minimum L2 distance ($d$)
from the distances between it and all the other keypoints in the same image,
then crop a $d \times d$ patch centered at this keypoint
and resize it to $224\times 224$~\footnote{We explain the approach of resizing patches to full image resolution instead of other methods to extract part representations in Appendix~\ref{part resize}}.
%
For each method, we directly use the pretrained encoders to extract features, without additional training processes, and take the better one from the class token or the average embedding of all patch tokens as the extracted representation. With each patch as the query patch, we attempt to retrieve patches belonging to the same category by calculating the cosine similarity between its representation and all the other patches' in the dataset and utilize the average precision (AP) as the retrieval metric. Finally, we average all the obtained AP scores (with all patches respectively taken as the query patch for retrieval) as the final retrieval score to evaluate the retrieval performance of the method.

The results are provided in Table~\ref{tab:quantitative_part_retrieval}.
We have the following observations.
\ding{182} Self-supervised models except BEiT and MAE outperform the supervised model DeiT, 
indicating the capability that contrastive learning and CAE learn part-aware representations.
BEiT and MAE perform inferior, consistent to the observations
in ImageNet classification in Table~\ref{tab:imagenet_classification}.
\ding{183} iBOT performs the best,
and the reason might be that
the capability of learning part-aware representations
is boosted
by making use of both contrastive learning
and masked image modeling. 

We also report the part retrieval performance
of the projected representations of contrastive learning methods in Table~\ref{tab:quantitative_part_retrieval}.
The performance is much lower than the encoded representations.
This provides an extra evidence for
the part-to-whole hypothesis of contrastive learning:
the projected representations are more about the whole object.

Finally, we visualize some patch retrieval results of the encoded representations on the ImageNet validation set in Figures~\ref{fig:visualization_1} (More visualization is presented in Appendix~\ref{app:visualization}). It is observed that the retrieved patches of self-supervised methods are generally more about the semantics of the query part than that of DeiT. 
The results demonstrate that the encoded representations of DeiT 
focus more on object-level semantics,
while the encoded representations of these self-supervised methods
are more about
part-level semantics.
Among these methods, the retrieved patches of MAE have less semantic correlation but often share similar hues.



\begin{figure}[t]
\footnotesize
    \centering
    \begin{overpic}[width=.98\linewidth]{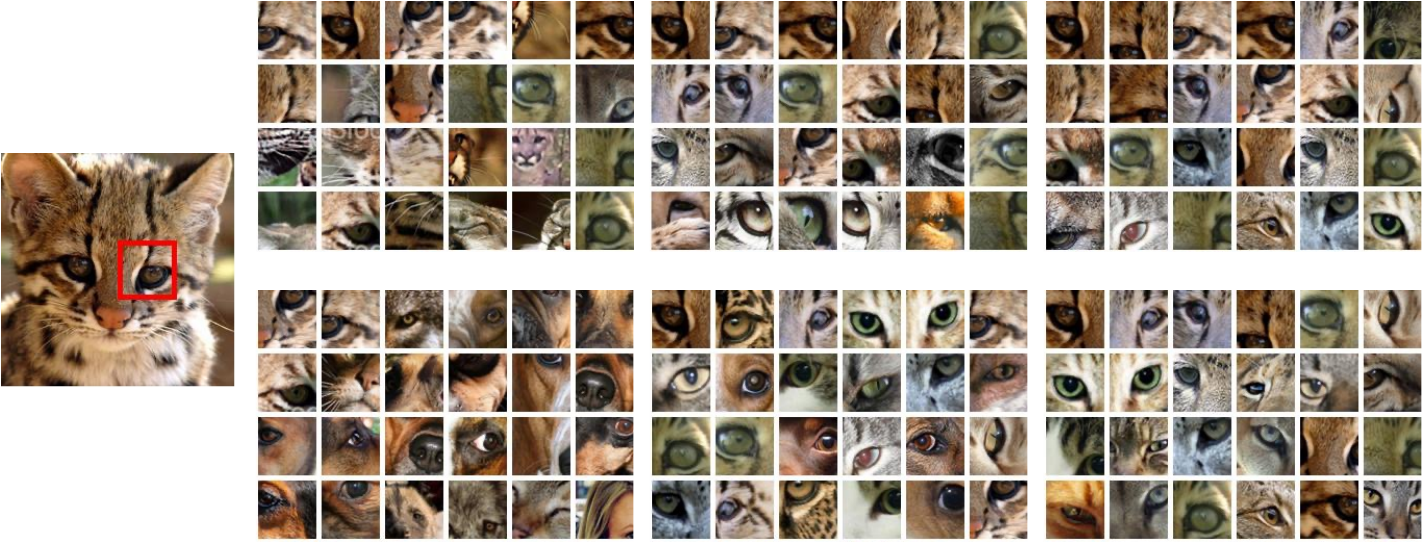}
    \put(28.7,38.4){DeiT}
    \put(54.5,38.4){MoCo v3}
    \put(83.8,38.4){DINO}
    \put(28.5,18.2){MAE}
    \put(56.6,18.2){CAE}
    \put(83.8,18.2){iBOT}
    \end{overpic}
    
    \vspace{0.5cm}

    \begin{overpic}[width=.98\linewidth]{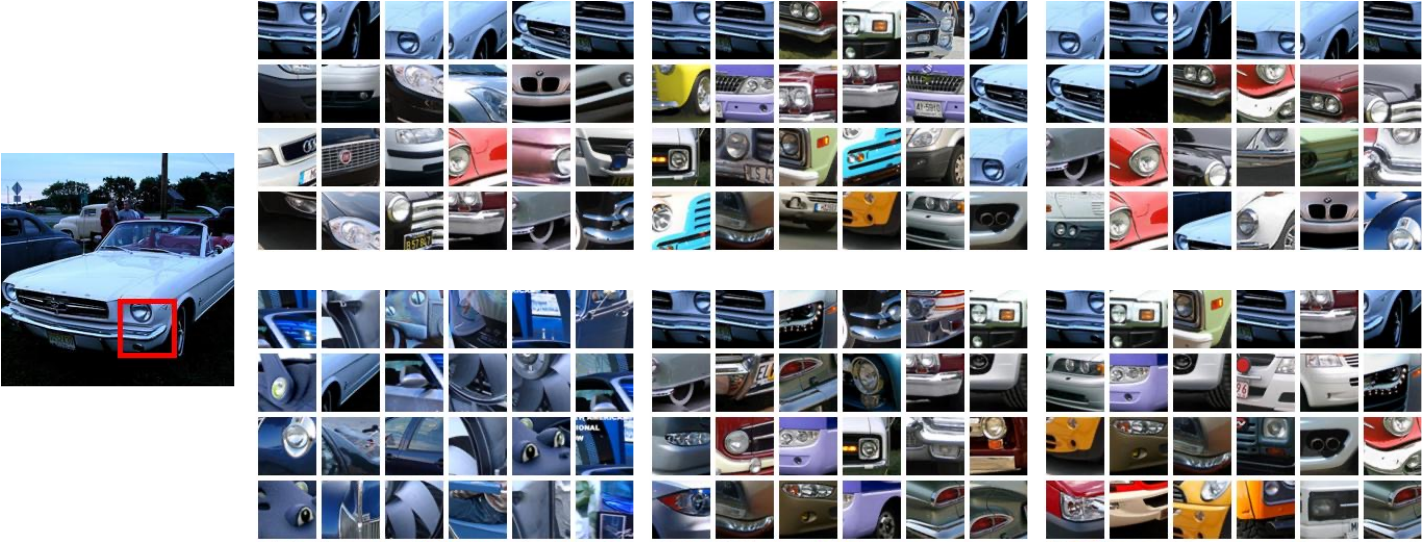}
    \put(28.7,38.4){DeiT}
    \put(54.5,38.4){MoCo v3}
    \put(83.8,38.4){DINO}
    \put(28.5,18.2){MAE}
    \put(56.6,18.2){CAE}
    \put(83.8,18.2){iBOT}
    \end{overpic}
    \caption{Patch retrieval comparisons of encoded representations on cropped patches from ImageNet.}
    \label{fig:visualization_1}
\end{figure}

\noindent\textbf{Part classification.}
We further conduct part classification experiments
on the datasets used for part retrieval, requiring pretrained models to assign the patches into the corresponding category label. 
We consider two kinds of extra learnable layers,
linear probing and attentive probing,
for classification. For linear probing, we learn a supervised linear classification layer on the extracted class token of the frozen encoders. While for attentive probing, following \cite{cae_chen2022context}, a cross attention module and a batch normalization layer without affine transformation are additionally inserted between the encoder and the linear classifier. And a new learnable class token is taken as the query of the cross attention module, to replace the original class token extracted by the frozen encoder. We use SGD optimizer with a learning rate of 0.4 and 0.04 for linear probing and attentive probing, respectively. For both linear probing and attentive probing, the models are trained for 90 epochs. And the momentum of SGD is set to 0.9, the weight decay is set to 0, and the batch size is set to 1024.

The results in Table~\ref{tab:quantitative_part_retrieval} show that:
\ding{182} While DeiT performs the best in the image classification task (see Table~\ref{tab:imagenet_classification}), for part classification, contrastive-based methods like MoCo v3, DINO, and iBOT outperform DeiT by more than 2\% under both linear and attentive probing settings. 
\ding{183} Though MIM-based models CAE and MAE are inferior to DeiT in object-level classification (\eg, more than 10\% and 4\% lower in linear and attentive probing), they show competitive performance in linear probing and higher results than DeiT in attentive probing, demonstrating they learn better part-aware representations.
\ding{184} BEiT is inferior to other works, and iBOT has good performance,
implying that the probing quality
of pretrained encoders
is a good indicator
for downstream performance.

\begin{table*}[t]
    \centering
    \footnotesize
    \caption{
    Part-level linear semantic segmentation results (\%) on the ADE20K-Part, Pascal-Part, and LIP datasets.
    }
    \label{tab:PASCALFCNSegmentation}
    \renewcommand{\arraystretch}{1.1}
    \setlength{\tabcolsep}{10pt}
\begin{tabular}{l c c c c c c c c  cccc}
    \shline
        \multirow{2}{*}{Methods} && \multicolumn{3}{c}{\tabincell{c}{ADE20K-Part \\ 209 Part Classes}} && \multicolumn{3}{c}{\tabincell{c}{Pascal-Part\\ 193 Part Classes}} && 
        \multicolumn{3}{c}{\tabincell{c}{LIP \\ 19 Part Classes}} \\
        \cline{3-5} \cline{7-9} \cline{11-13} 
        && mIoU & mAcc & aAcc && mIoU & mAcc & aAcc && mIoU & mAcc & aAcc\\
        \shline
        \multicolumn{4}{l}{\emph{Supervised Model}:}\\
        DeiT   && 27.3 & 34.7 & 69.2  && 27.4 & 36.1 & 65.8 && 41.4 & 52.6 & 73.5  \\
        \hline
      \multicolumn{4}{l}{\emph{Contrastive Learning}:}\\
        MoCo v3  && 27.1 & 34.7 & 70.1 && 27.1 & 35.8 & 66.0 && 41.9 & 53.0 & 74.5\\
        DINO && 28.9 & 36.8 & 70.3  && 27.8 & 36.5 & 66.4 && 41.0 & 51.9 & 74.0 \\
        \hline
      \multicolumn{4}{l}{\emph{Masked Image Modeling (MIM)}:}\\
        BEiT  && 18.6 & 25.8 & 58.2 && 14.8 & 21.4 & 47.0 && 27.2 & 36.5 &  60.1 \\
        MAE  && 26.3 & 35.0 & 67.3 &&  24.3 & 32.9 & 61.5 && 38.2 & 48.7 & 71.3 \\
        CAE  && 28.4 & 36.9 & 71.1 && 27.8 & 37.0 & 66.3 && 43.7 & 55.1 & 75.9 \\
        \hline
      \multicolumn{4}{l}{\emph{Contrastive Learning + MIM}:}\\
        iBOT && 32.2 & 40.0 & 73.4 && 30.7 & 40.0 & 69.7 && 44.6 & 55.7 & 76.6\\
        \shline
    \end{tabular}
    \vskip -0.12in
\end{table*}

\noindent\textbf{Part segmentation.}
We perform part-level linear semantic segmentation to study the finer-grained part representation modeling capability of different pretraining paradigms on three widely used datasets: 
ADE20K-Part~\citep{ade20k_zhou2019semantic} containing 209 parts from the ADE20K dataset~\citep{ade20k_zhou2019semantic, zhu2023crf},
Pascal-Part~\citep{pascal_part_chen2014detect} including 193 part categories,
and LIP~\citep{LIP_Gong_2017_CVPR} consisting of 19 semantic human part labels. See Appendix~\ref{app:datasets} for more dataset details.
%
Similar to the object-level semantic segmentation experiments, linear evaluation is employed here.
We maintain the same training protocols including learning rate ($4e-4$), training iterations ($160k$), and batch size ($16$) for all methods for fair comparisons~\footnote{We also consider to do some hyperparameter tuning per method for more fair comparisons. See Appendix~\ref{learning rate}}.  For ADE20K, the input size is set to $512 \times 512$ following previous works~\citep{bao2021beit, he2021masked, cae_chen2022context, zhou2021ibot}. For Pascal-Part, we adopt  $480 \times 480$ as image input resolution following~\cite{mmseg2020}. As for LIP, we use the same input size ($320 \times 320$) proposed in LIP~\citep{LIP_Gong_2017_CVPR}.

\begin{figure*}[t]
\centering
\includegraphics[width=1\linewidth]{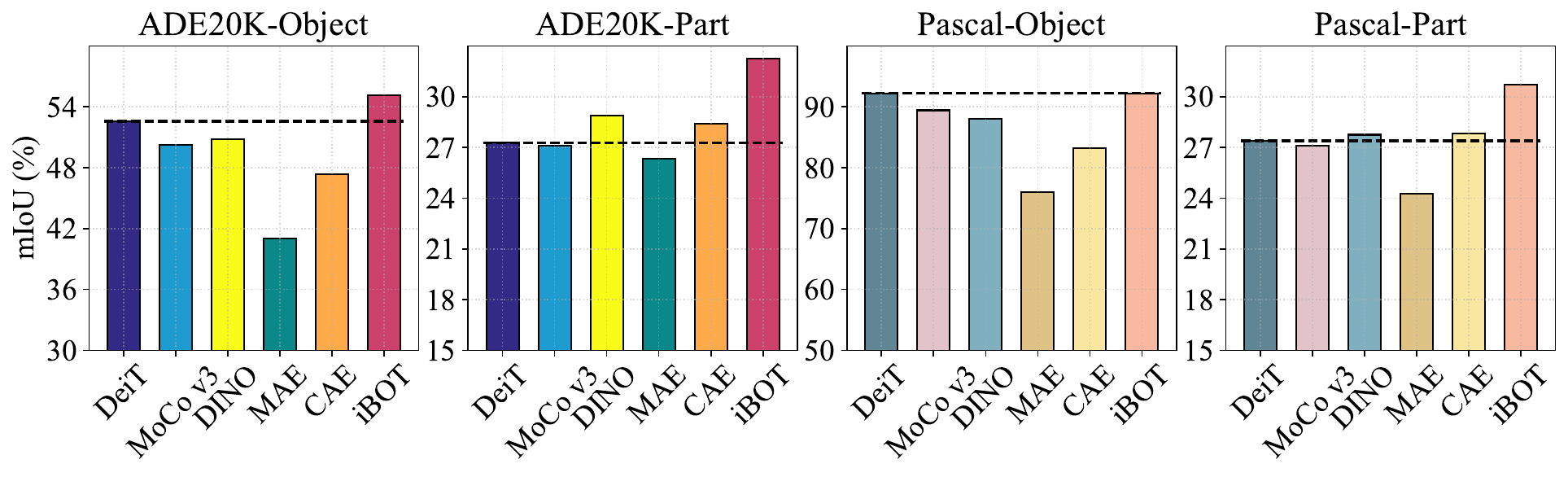}
\vspace{-12pt}
\vskip -0.12in
\caption{Comparisons between object-level and part-level semantic segmentation on ADE20K and Pascal-Part datasets.
Though the supervised DeiT is superior over self-supervised models (\ie, MoCo v3, DINO, MAE, CAE) on object-level segmentation, 
it is generally inferior to self-supervised models on part segmentation, demonstrating self-supervised methods learn good part-aware representations. 
iBOT enjoys the benefits of contrastive learning and MIM. See Appendix~\ref{app:detailed_segmentation_results} for detailed results.
}
\label{fig:object_part_comparison}
\vskip -0.1in
\end{figure*}

The results are reported in Table~\ref{tab:PASCALFCNSegmentation} with the following observations. 
\ding{182} Contrastive learning models, \ie, MoCo v3 and DINO, achieve competitive performance with the supervised model DeiT: DINO outperforms DeiT on ADE20K-Part and Pascal-Part, and MoCo v3 outperforms DeiT on LIP.
\ding{183} The MIM model CAE, outperforms DeiT by large margins on all three datasets, \eg, 1.1\% on ADE20K-Part and 2.3\% on LIP, indicating CAE learns good part-aware representations.
Similar to part retrieval, BEiT and MAE perform inferior in absolute metrics possibly due to pretraining quality in representation encoding~\footnote{We discuss the reasons in detail and add human pose estimation experiments in Appendix~\ref{discussion}}.
Also, it is worth mentioning that the performance gap (1.0\%) between DeiT and MAE on part-level segmentation is significantly narrowed compared to that (11.6\%) of object-level segmentation as shown in Figure~\ref{fig:object_part_comparison}. This indicates that although MAE does not outperform DeiT on part-level tasks,  it tends to learn more about parts than DeiT. 
Similar observations on BEiT can also be found in Appendix~\ref{app:detailed_segmentation_results}.
\ding{184} Compared with object-level segmentation results in Table~\ref{tab:ADE20K_segmentation},
DeiT learns better object-level semantics by explicit supervision than both contrastive learning and MIM, however, it is generally inferior to self-supervised models on part segmentation.
\ding{185} The model iBOT, which leverages both contrastive learning and MIM, outperforms all other works on three datasets, demonstrating its powerful capability in learning finer part-level semantics. Combining the two self-supervised learning techniques is thus a promising direction.

\textit{In summary, we show that self-supervised methods are potentially capable of learning part-aware representations.}
Among them, CAE is a representative MIM work, showing good performance by explicitly predicting the encoded representations of the masked patches in the encoding space;
contrastive learning methods MoCo v3 and DINO outperform BEiT and MAE;
and iBOT performs the best~\footnote{The iBOT authors verified that introducing MIM brings richer semantics to the patch tokens in Sec.~4.3.1 of their paper. 
This could be the reason for the superior performance of object-level and part-level tasks observed in iBOT where MIM and CL are combined.
}.
The observations are evidenced by three part-based segmentation benchmarks consistently.

\subsection{Observation Summary between Object-Level and Part-Level Segmentation}
We conduct both object-level and part-level linear semantic segmentation on different hierarchies of the same dataset.
Considering that the 209 classes in ADE20K-Part are basically chosen from 59 object classes, we denote the 59-object dataset as ADE20K-Object. Similarly, Pascal-Object consists of 16 object categories, corresponding to the 193 part categories in Pascal-Part.

The results in Figure~\ref{fig:object_part_comparison} show that:  
although self-supervised models (except iBOT) are inferior to the supervised DeiT on ADE20K-Object and Pascal-Object, they significantly narrow the performance gap on ADE20K-Part and Pascal-Part and even outperform DeiT, demonstrating self-supervised methods tend to learn more about parts and potentially produce good part-aware representations.
Similar observations could be found from the object classification in Table~\ref{tab:imagenet_classification} and part classification in Table~\ref{tab:quantitative_part_retrieval}.

In comparison to contrastive learning, CAE shows a stronger capability of learning part-aware representations, and a weaker capability of learning object-level semantics.
The superiority of iBOT,
a combination of contrastive learning and masked image modeling,
demonstrates that
it enjoys the benefits of contrastive learning and masked image modeling. 
We have outlined several key takeaways in Appendix~\ref{app:takeaways} derived from our experiments and analysis, which we hope could inspire further works in self-supervised learning.

Additionally, although our main focus is not on in-the-wild datasets like COCO and OpenImages, we conduct a preliminary exploration on them. 
Following the same setting as in our paper, we conducted a quick experiment with a transformer-based Long-seq-MAE~\citep{hu2022exploring} that is pretrained on COCO. Our experiments show that its part retrieval (32.3\% AP) on CUB-200 and LIP segmentation results (45.89\% mIoU) outperform MAE (28.5\% AP and 38.2\% mIoU) pretrained on ImageNet1k. It also narrows the gap between DeiT (35.0\% AP) in part retrieval and outperforms DeiT (41.4\% mIoU) in part segmentation. These results indicate that pretraining on more in-the-wild datasets has the potential to learn better part representations as cropping can produce more parts of multiple objects in one image. We leave it as an interesting future direction.

\section{Conclusion}
We attempt to study the capability of learning part-aware representations
of self-supervised representation pretraining methods.
We provide speculations for contrastive learning
and masked image modeling:
part-to-whole and part-to-part,
with empirical results 
justifying the speculations. 
Our study presents an aspect to understand
what self-supervised representation pretraining methods
learn.

\textbf{Limitation and future work.} The part-aware representation learning capability is one of the properties of self-supervised pretraining. There should be other characteristics that are worth exploring.
The proposed approach has no ethical or societal issues on its own, except those inherited from computer vision.

\section*{Acknowledgments}
This work is supported by National Science Foundation of China (NSFC) Grant No. 61972008. 

\bibliography{main}
\bibliographystyle{tmlr}

\appendix
\newpage
\section{Appendix}


\subsection{Model Description}
\label{app:model_description}

For all the models involved in the experiments including DeiT~\citep{touvron2020deit}, MoCo v3~\citep{mocov3_chen2021empirical}, DINO~\citep{caron2021emerging}, BEiT~\citep{bao2021beit}, MAE~\citep{he2021masked}, CAE~\citep{cae_chen2022context}, and iBOT~\citep{zhou2021ibot}, we use their official code to implement the encoders.
It is worth noticing that for DINO and iBOT, we choose the checkpoint of the teacher models as they have been reported to perform better than the student models in their papers~\citep{caron2021emerging, zhou2021ibot}.

\subsection{Datasets}
\label{app:datasets}
\noindent\textbf{ADE20K}~\citep{ade20k_zhou2019semantic} is one of the most challenging benchmarks, containing 150 fine-grained semantic concepts and a variety of scenes with 1,038 image-level labels. There are 20,210 images in the training set and 2,000 images in the validation set. We choose 59 out of total 150 semantic concepts that are concrete objects containing parts~\citep{ade20k_zhou2019semantic}, termed ADE20K-Object. We also select 209 part categories that emerge both in the training set and the validation set, called ADE20K-Part.

\noindent\textbf{Pascal-Part}~\citep{pascal_part_chen2014detect} is a set of additional annotations for PASCAL VOC 2010~\citep{pascal-voc-2010}, thereby holding the same statistics as those of PASCAL VOC 2010. It provides segmentation masks for each part of objects. Concretely, the dataset includes 20 object-level categories and 193 part-level categories. In our experiments, we remove 4 object categories that do not contain parts including boat, table, chair, and sofa.

\noindent\textbf{LIP}~\citep{LIP_Gong_2017_CVPR} is a large-scale benchmark for human parsing research, which includes 50,462 images with pixel-wise annotations on 19 semantic part labels. In detail, it includes 19,081 full-body images, 13,672 upper-body images, 403 lower-body images, 3,386 head-missed images, 2,778 back-view images and 21,028 images with occlusions. There are 30,462 images in the training set and 10,000 images in the validation set. The rest 10,000 images are served as the test set with missing labels for competition evaluation.

\noindent\textbf{CUB-200-2011}~\citep{cub200_wah2011caltech} is a popular benchmark for fine-grained image classification, and also provides bounding box and part location annotations. It contains 11,788 images of 200 bird species and 15 part keypoint annotations per bird. In this work, we mainly leverage its part keypoint annotations. And only 4 part categories (right eye, right leg, left wing, and tail) are chosen to be considered in our experiments, to make sure that the selected keypoints are far enough away from each other and enough context information can be contained in the cropped patches. (We also tried using all keypoints and the conclusion is consistent.)

\noindent\textbf{COCO}~\citep{caesar2018coco}, as one of the most widely-used human pose estimation datasets, contains more than 200,000 images and 250,000 labeled person instances. Similar to CUB-200-2011 mentioned above, only 3 (nose, right wrist, and left ankle) of its 17 keypoint categories are considered in our experiments.

\noindent \textbf{CIFAR10}~\citep{krizhevsky2009learning} is a widely-used dataset that has $50,000$ training images and $10,000$ test images. CIFAR10 has $10$ categories. The dimension for CIFAR10 images is $32 \times 32 \times 3$.

\subsection{Part Resizing Approach}\label{part resize}
We explain the idea behind our part resizing approach below. There are three main reasons for using this approach:

$\bullet$ To maintain the shape of the position embedding the same as during pretraining, it is essential for the transformer to distinguish patches located at different positions. Rashly resizing position embedding, \eg, through interpolation, may introduce some noise and cause a mismatch with the parameters of other pretrained transformer components.

$\bullet$ Is there a domain gap between training data and the resized part? No may be the answer.  As shown in Table~\ref{tab:crop scale}, the crop operation during self-supervised pretraining, mostly adopts a scale range of $(0.08-1)$, followed by image resizing. Hence, the pretrained models may see these "upscaled" parts during training potentially. Therefore, there is no domain gap using our part resizing approach.

\begin{table}[h]
    \centering
    \footnotesize
    \setlength{\tabcolsep}{12pt}
    \renewcommand{\arraystretch}{1.2}
    \caption{The crop scale of different pretrained models.
    }
    \label{tab:crop scale}
\begin{tabular}{l c c c c c c c}
    \shline
        Method & DeiT &	MoCo v3 & DINO & BEiT &	MAE	& CAE & iBOT \\
        \shline
        Crop Scale & 0.08-1 & 0.08-1 & 0.4-1, 0.05-0.4 & 0.08-1 & 0.2-1 & 0.08-1 & 0.14-1, 0.05-0.4\\
        \shline
    \end{tabular}
\end{table}  

$\bullet$ We also considered another manner: processing the whole image and then taking the individual patch representations. However, due to the presence of the attention module, different patch representations would interact globally. Therefore, the part representation would contain information from other parts and may lead to information leaks hence the confusion about part representations.

\subsection{Detailed Results for Object-Level and Part-Level Segmentation}
\label{app:detailed_segmentation_results}
In this section, we provide detailed comparisons between object-level and part-level semantic segmentation in Table~\ref{tab:ADE20KSegmentation} and Table~\ref{tab:PASCALSegmentation}.
Similar observations as in Figure~\ref{fig:object_part_comparison} in the main paper are found: although the supervised DeiT is superior over self-supervised methods on ADE20K-Object and Pascal-Object except iBOT, it is generally inferior to self-supervised models on ADE20K-Part and Pascal-Part, demonstrating self-supervised methods can learn good part-aware representations. 
BEiT and MAE perform inferior, perhaps because the two methods do not have an explicit process to predict the encoded representations of masked patches, instead, directly reconstruct the targets.

\begin{table}[h]
    \centering
    \footnotesize
    \caption{
    Linear semantic segmentation results on ADE20K-Object and ADE20K-Part.
    }
    \vspace{-4pt}
    \label{tab:ADE20KSegmentation}
    \renewcommand{\arraystretch}{1.2}
    \setlength{\tabcolsep}{9pt}
\begin{tabular}{l c c c c c c c c}
    \shline
        \multirow{2}{*}{Methods} & & \multicolumn{3}{c}{\tabincell{c}{Object Seg on ADE20K-Object \\ 59 Object Classes}} & & \multicolumn{3}{c}{\tabincell{c}{Part Seg on ADE20K-Part \\ 209 Part Classes}} \\ \cline{3-5} \cline{7-9}
        && mIoU & mAcc & aAcc & & mIoU & mAcc & aAcc \\
        \shline
        \multicolumn{4}{l}{\emph{Supervised Model}:} \\
        DeiT   && 52.6 & 62.9 & 83.8 && 27.3 & 34.7 & 69.2  \\ 
        \hline
      \multicolumn{4}{l}{\emph{Contrastive Learning}:}\\
        MoCo v3  && 50.2 & 60.4 & 83.6 && 27.1 & 34.7 & 70.1  \\ 
        DINO && 50.8 & 60.8 & 83.9  && 28.9 & 36.8 & 70.3 \\  
        \hline
      \multicolumn{4}{l}{\emph{Masked Image Modeling (MIM)}:}\\
        BEiT  && 28.6 & 37.2 & 73.4 && 18.6 & 25.8 & 58.2 \\ 
        MAE  && 41.0 & 50.6 & 79.9 && 26.3 & 35.0 & 67.3  \\ 
        CAE  && 47.4 & 58.4 & 82.9 && 28.4 & 36.9 & 71.1 \\ 
        \hline
      \multicolumn{4}{l}{\emph{Contrastive Learning + MIM}:}\\
        iBOT && 55.2 & 65.1 & 85.6 && 32.2 & 40.0 & 73.4 \\ 
        \shline
    \end{tabular}
\end{table}

\begin{table}[h]
    \centering
    \footnotesize
    \caption{
    Linear semantic segmentation results on Pascal-Object and Pascal-Part.
    }
    \vspace{-4pt}
    \label{tab:PASCALSegmentation}
    \renewcommand{\arraystretch}{1.2}
    \setlength{\tabcolsep}{9.6pt}
\begin{tabular}{l c c c c c c c c }
    \shline
        \multirow{2}{*}{Methods} && \multicolumn{3}{c}{\tabincell{c}{Object Seg on Pascal-Object \\ 16 Object Classes}} && \multicolumn{3}{c}{\tabincell{c}{Part Seg on Pascal-Part\\ 193 Part Classes}} \\
        \cline{3-5} \cline{7-9} 
        && mIoU & mAcc & aAcc && mIoU & mAcc & aAcc \\
        \shline
        \multicolumn{4}{l}{\emph{Supervised Model}:}\\
        DeiT   && 92.2 & 95.3 & 96.8  && 27.4 & 36.2 & 65.8  \\
        \hline
      \multicolumn{4}{l}{\emph{Contrastive Learning}:}\\
        MoCo v3  && 89.4 & 93.7 & 95.7 && 27.1 & 35.8 & 66.0  \\
        DINO && 88.0 & 92.7 & 95.3  && 27.8 & 36.5 & 66.4  \\
        \hline
      \multicolumn{4}{l}{\emph{Masked Image Modeling (MIM)}:}\\
        BEiT  && 56.4 & 69.0 & 76.8 && 14.8 & 21.4 & 47.0 \\
        MAE  && 76.1 & 84.6 & 89.5 && 24.3 & 32.9 & 61.5  \\
        CAE  && 83.3 & 89.7 & 93.2 && 27.8 & 37.0 & 66.3  \\
        \hline
      \multicolumn{4}{l}{\emph{Contrastive Learning + MIM}:}\\
        iBOT && 92.1 & 95.3 & 97.1 && 30.7 & 40.0 & 69.7 \\
        \shline
    \end{tabular}
\end{table}

\subsection{Ablation on Learning Rate for Part Semantic Segmentation}\label{learning rate}

We use batch size 16 and 160k iterations following previous methods~\cite{he2021masked, bao2021beit, zhou2021ibot}. For optimizer, we adopt AdamW which is widely used for Vision Transformer~\citep{DosovitskiyB0WZ21}. As for learning rate, we conduct experiments on LIP part segmentation with different learning rates including $1e-5$, $2e-4$, $3e-4$, $4e-4$, $5e-4$, and $1e-3$ for all pretrained models. The results are presented in Table~\ref{tab:learn_rate}. The best results are bold. One can see that except for $1e-5$, other learning rates slightly influence the final performance. In terms of the overall performance of each model, MoCo v3, CAE, and iBOT show superior performance over supervised model DeiT, validating our findings in the main paper. 

\subsection{Why BEiT and MAE Perform Inferior?} 
\label{discussion}
In linear evaluation, the quality of output representations always depends on the pretraining quality of the encoder. A well-trained encoder usually produces representations of high quality. In our experiments, we find that BEiT only obtains 56.7 linear probe on ImageNet while other models achieve about 70 linear probe on ImageNet. \textbf{The large gap indicates that BEiT does not learn high-quality representations as other models, \ie, its pretraining quality is not good enough.} This could be the main reason that it performs worst on downstream tasks (both object-level and part-level tasks).
 
For MAE, we find that the colors of the retrieved patches in Figure~\ref{fig:visualization_1} and Figure~\ref{fig:visualization_2} in Appendix are quite similar. \textbf{Based on it, we speculate that its low-level color reconstruction target helps MAE learn better middle- or low-level features while providing limited high-level semantics.} To validate this point, we further conduct experiments on the human pose estimation task, which has been proven by previous work~\cite{ding2021hr} to be a task requiring middle-level features. Similar to linear probing, we train one linear layer over the features extracted by the frozen backbones to predict human keypoints. The results are shown in Table~\ref{tab:humanpose} and MAE outperforms all other counterparts by large margins. So we believe MAE tends to learn lower-level features and pay relatively less attention to the high-level semantics, which leads to its inferior performance on classification and segmentation tasks.

\begin{figure}[t]
\begin{minipage}[t]{0.51\linewidth}
\begin{table}[H]
\centering  
  \caption{
  Different learning rates in LIP part segmentation.
  }
  \setlength{\tabcolsep}{5pt}
\renewcommand{\arraystretch}{1.2}
    \footnotesize 
    \begin{tabular}{l c c c c c c}
      \shline
      Method & $1e-5$ & $2e-4$ & $3e-4$ & $4e-4$ & $5e-4$ & $1e-3$ \\
      \shline
      \multicolumn{5}{l}{\emph{Supervised Model}:}\\
      DeiT & 40.5 &	41.3 &	41.4 &	\textbf{41.4} & 41.3 &	41.3 \\
      \hline
      \multicolumn{5}{l}{\emph{Contrastive Learning}:}\\
      MoCo v3 & 40.0 & 41.7 &	41.7 & \textbf{41.9} &	41.7 &	41.8\\
      DINO & 38.1 & 40.8 & 40.9 & \textbf{41.0} & 40.9 & 40.9 
      \\
      \hline
      \multicolumn{5}{l}{\emph{Masked Image Modeling (MIM)}:}\\
      BEiT  & 23.1	& 27.2 & 27.4 & 27.2 & 27.1 & \textbf{27.5} \\
      MAE  & 34.7 &	38.5 &	38.6 & 38.2 & 38.7 & \textbf{38.8} \\
      CAE  & 42.8 & 44.1 & 44.2 & 43.7 & \textbf{44.3}	& 44.3 \\
      \hline
      \multicolumn{5}{l}{\emph{Contrastive Learning + MIM}:}\\
      iBOT  & 41.5 & 44.4 & 44.5 & 44.6 & 44.5 & \textbf{44.8} \\
      \shline
  \end{tabular}
  \label{tab:learn_rate}
\end{table}
\end{minipage}
\hfill
\begin{minipage}[t]{0.45\linewidth}
\begin{table}[H] 
    \centering
    \footnotesize
    \renewcommand{\arraystretch}{1.2}
    \setlength{\tabcolsep}{9pt}
    \caption{Linear human pose estimation results on the COCO dataset.}
    \vspace{0.2cm}
    \begin{tabular}{l c c c c}
    \shline
         Method & AP & AP50 & AR & AR50 \\
        \shline
        \multicolumn{4}{l}{\emph{Supervised Model}:}\\
        DeiT & 6.8 &	27.5	&14.7&	44.3 \\
        \hline
      \multicolumn{4}{l}{\emph{Contrastive Learning}:}\\
        MoCo v3 & 12.6	& 42.1& 21.61	& 56.3  \\
        DINO & 16.6 &	49.0	&25.9	&62.1  \\
        \hline
      \multicolumn{4}{l}{\emph{Masked Image Modeling (MIM)}:}\\
        MAE & \textbf{23.6} &	\textbf{59.8}	&\textbf{33.3}	&\textbf{69.6}  \\
        CAE  &  17.5 & 	50.1	 & 28.9  & 	64.0  \\
        \hline
      \multicolumn{4}{l}{\emph{Contrastive Learning + MIM}:}\\
        iBOT & 15.5	& 48.1 &	26.0 & 62.8 \\
        \shline
    \end{tabular}
    \label{tab:humanpose}
\end{table}
\end{minipage}
\end{figure}

\subsection{Takeaways}\label{app:takeaways}
 We have outlined several key takeaways derived from our experiments and analysis, which we believe could benefit further works in the self-supervised learning field.

\begin{itemize}
\setlength\itemsep{0.2 em}
    \vspace{-8pt}
    \item \textbf{1)} Supervised models care more about the whole objects, while self-supervised models have a stronger capability of learning part-aware representations.
    understanding self-supervised representation pretraining.
    \item \textbf{2)} Contrastive learning may learn higher-level semantics and be more semantically abundant than MIM. Combining MIM and CL is potentially capable of learning multi-level semantics.
    \item \textbf{3)} MIM methods, e.g., CAE, learn slightly better part-level features than contrastive learning methods. While the learned features of MIM methods, to some extent, depend on the reconstruction targets, e.g., the RGB (color) target used in MAE leads it to learn lower-level features, hence only sub-optimal results on high-level tasks.
\end{itemize}

Those takeaways are evidenced correspondingly as follows:
\begin{itemize}
\setlength\itemsep{0.2em}
    \vspace{-8pt}
    \item \textbf{1)} In Figure~\ref{fig:object_part_comparison}, Table~\ref{tab:ADE20KSegmentation}, and Table~\ref{tab:PASCALSegmentation}, from object-level to part-level, considerable performance improvements (relative to DeiT) are observed for all self-supervised models. Among these methods, DINO, CAE, and iBOT show larger improvements, demonstrating they can learn better features for part-aware segmentation. For example, DeiT (52.6\% in mIoU) outperforms DINO (50.8\%) and CAE (47.4\%) by 1.8\% and 5.2\% in ADE20K object-level experiment. However, in ADE20K part-level experiment, the situation is completely opposite. DINO (28.9\% in mIoU) outperforms DeiT (27.3\%) by 1.6\% and CAE (28.4\%) outperforms DeiT by 1.1\%. Similar results can be seen in segmentation experiments on the Pascal dataset, as well as classification results on COCO and CUB-200.
    \item \textbf{2)} As shown in the second row of Figure~\ref{fig:searchresultsteaser}, the retrieved patches of MoCo v3 are all about dog mouths. While the retrieved patches of CAE contain some patches about the mouths of cats or foxes. Similarly, in Figure~\ref{fig:visualization_2}, the retrieved patches of CAE for the watch also include patches about the dial of the dial phone. Contrastive learning methods are more likely to retrieve patches with the same category while CAE sometimes retrieves some patches with similar textures. This observation will be more frequently found from the top-100 retrieved patches, based on which we show contrastive learning methods tend to learn better high-level semantics than MIM. And as shown in our paper, iBOT outperforms other methods by large margins in almost all tasks, including object-level and part-level tasks.
    \item \textbf{3)} \ding{182} On part retrieval, CAE outperforms contrastive learning methods by a large margin (about 8\% and 5\% on CUB-200 and COCO, respectively). On object-level linear probing or segmentation tasks, CAE performs clearly worse than contrastive learning methods. While this performance gap is significantly narrowed even closed on corresponding part-level tasks. \ding{183} the retrieved patches of MAE in Figure~\ref{fig:visualization_1} and Figure~\ref{fig:visualization_2} tend to share similar hues, and MAE performs much better than other methods on the human pose estimation task which requires middle-level features.
\end{itemize}

\subsection{More Discussion about Related Work}\label{app:discussion}
Image crop is important in self-supervised learning and brings promising properties to self-supervised models: ~\cite{van2021revisiting} shows that MoCo learns spatially structured representations when trained with a multi-crop strategy; By leveraging object-aware cropping, ~\cite{mishra2021object} encourages the self-supervised model to learn both object and scene level semantic representations from real-world uncurated datasets. Similarly, in our part-to-whole explanation, random crop on ImageNet generates views that describe parts of an object and are encoded to part representations and projected to whole for agreement. DenseCL~\citep{wang2021dense} designs a dense self-supervised learning method that directly works at the level of pixels (or local features) by taking into account the correspondence between local features. DetCo~\citep{xie2021detco} achieves better performance trade-off on both classification and detection through multi-level supervision to intermediate representations and contrastive learning between global image and local patches.

\subsection{More Examples for ImageNet Patch Retrieval Visualization}
\label{app:visualization}

\begin{figure}
\footnotesize
    \centering
    
    \begin{overpic}[width=.98\linewidth]{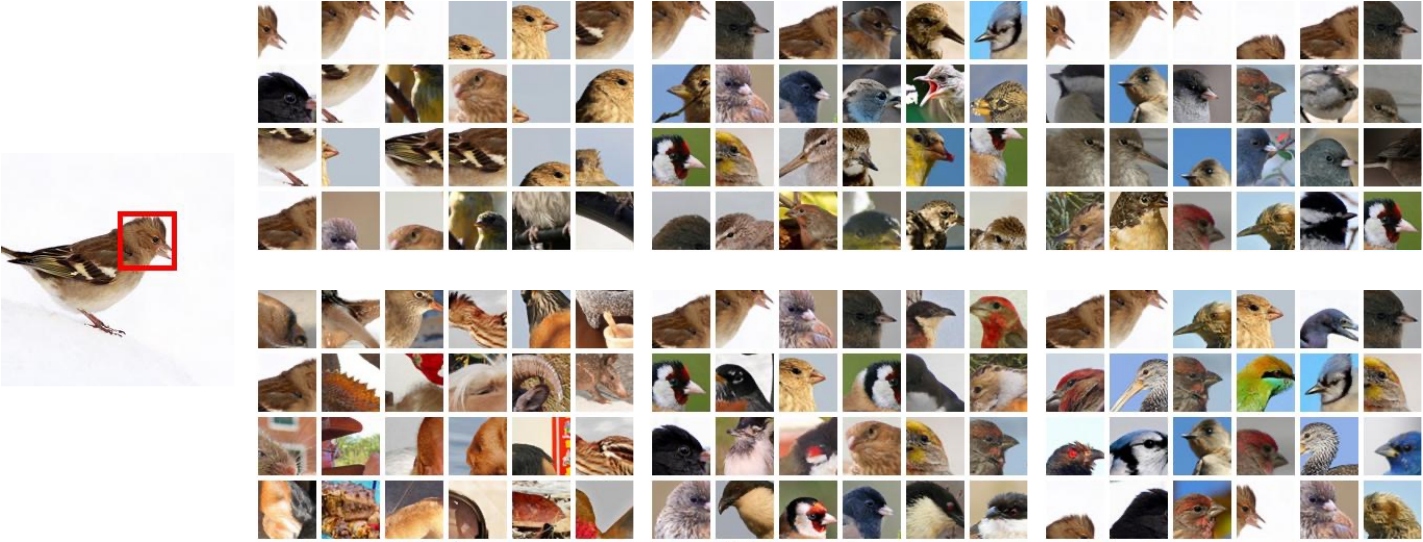}
    \put(28.7,38.4){DeiT}
    \put(54.5,38.4){MoCo v3}
    \put(83.8,38.4){DINO}
    \put(28.5,18.2){MAE}
    \put(56.6,18.2){CAE}
    \put(83.8,18.2){iBOT}
    \end{overpic}
    

    \vspace{0.5cm}

    \begin{overpic}[width=.98\linewidth]{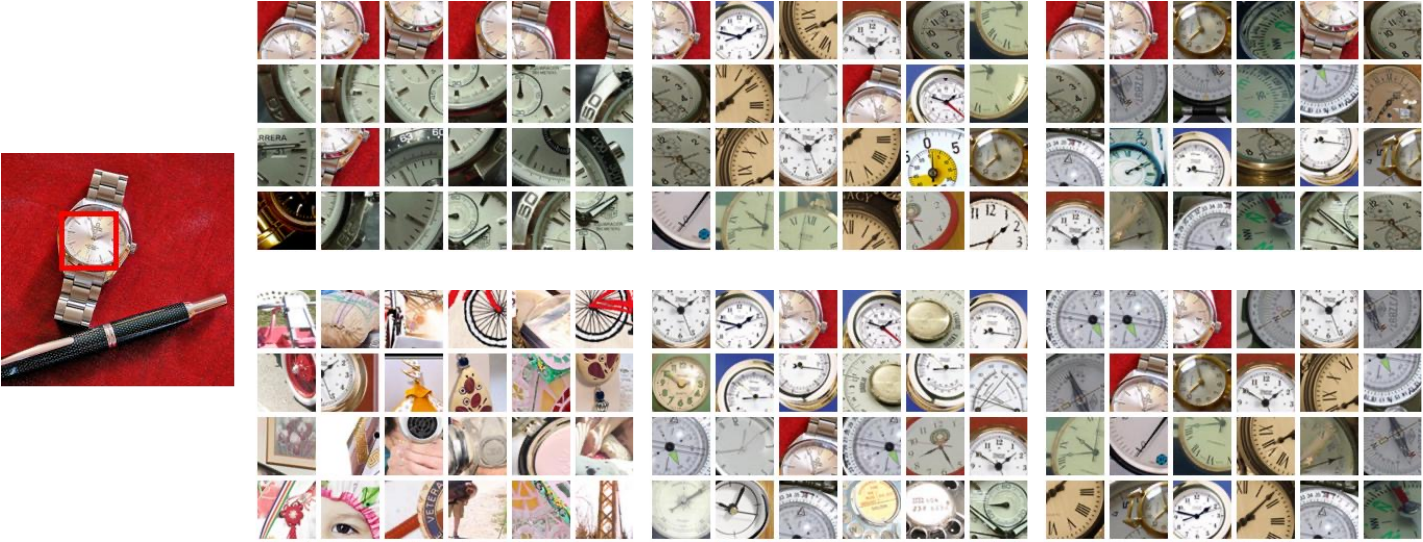}
    \put(28.7,38.4){DeiT}
    \put(54.5,38.4){MoCo v3}
    \put(83.8,38.4){DINO}
    \put(28.5,18.2){MAE}
    \put(56.6,18.2){CAE}
    \put(83.8,18.2){iBOT}
    \end{overpic}

    \vspace{0.5cm}

    \begin{overpic}[width=.98\linewidth]{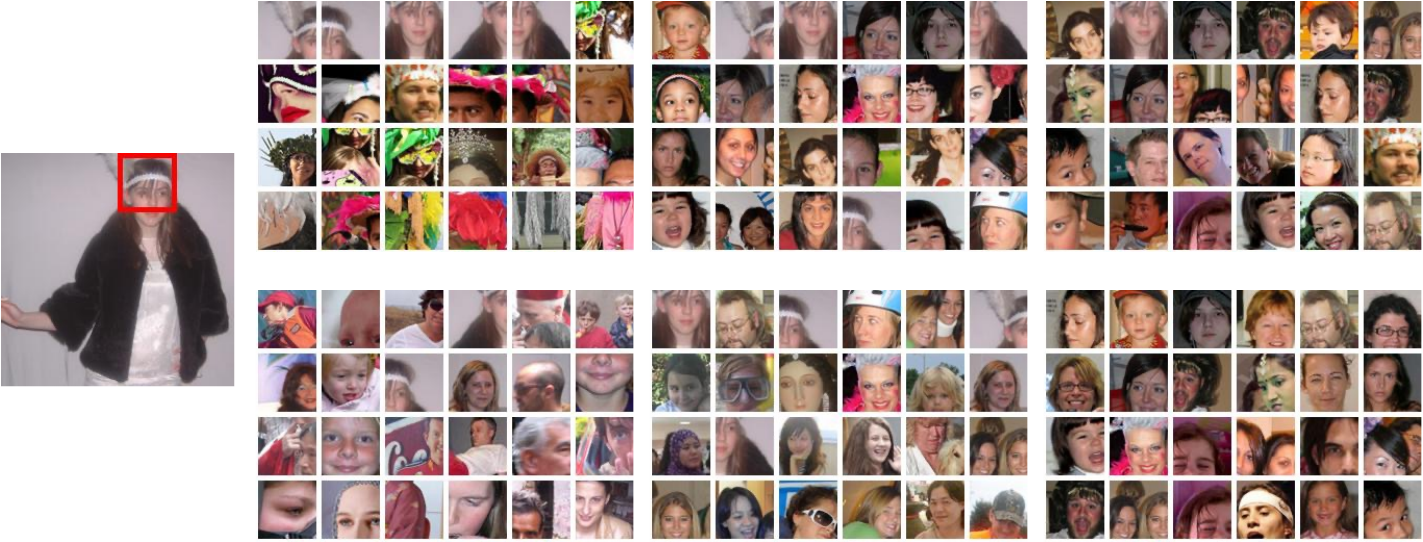}
    \put(28.7,38.4){DeiT}
    \put(54.5,38.4){MoCo v3}
    \put(83.8,38.4){DINO}
    \put(28.5,18.2){MAE}
    \put(56.6,18.2){CAE}
    \put(83.8,18.2){iBOT}
    \end{overpic}
    \caption{Patch retrieval comparisons of encoded representations on cropped patches from ImageNet.}
    \label{fig:visualization_2}
\end{figure}

We visualize more patch retrieval results of the encoded representations on the ImageNet validation set in Figures~\ref{fig:visualization_2}.

\end{document}